# Reasoning about Minimal Belief and Negation as Failure

**Riccardo Rosati**                                         ROSATI@DIS.UNIROMA1.IT
*Dipartimento di Informatica e Sistemistica*
*Università di Roma "La Sapienza"*
*Via Salaria 113, 00198 Roma, Italy*

## Abstract

We investigate the problem of reasoning in the propositional fragment of MBNF, the logic of minimal belief and negation as failure introduced by Lifschitz, which can be considered as a unifying framework for several nonmonotonic formalisms, including default logic, autoepistemic logic, circumscription, epistemic queries, and logic programming. We characterize the complexity and provide algorithms for reasoning in propositional MBNF. In particular, we show that skeptical entailment in propositional MBNF is $\Pi_3^p$-complete, hence it is harder than reasoning in all the above mentioned propositional formalisms for nonmonotonic reasoning. We also prove the exact correspondence between negation as failure in MBNF and negative introspection in Moore's autoepistemic logic.

## 1. Introduction

Research in the formalization of commonsense reasoning has pointed out the need of formalizing agents able to reason introspectively about their own knowledge and ignorance (Moore, 1985; Levesque, 1990). Modal epistemic logics have thus been proposed, in which modalities are interpreted in terms of knowledge or belief. Generally speaking, the conclusions an introspective agent is able to draw depend on both what she knows and what she *does not* know. Hence, any such conclusion may be retracted when new facts are added to the agent's knowledge. For this reason, many *nonmonotonic* modal formalisms have been proposed in order to characterize the reasoning abilities of an introspective agent.

Among the nonmonotonic modal logics proposed in the literature, the logic of minimal belief and negation as failure MBNF (Lifschitz, 1991, 1994) is one of the most studied formalisms (Chen, 1994; Bochman, 1995; Beringer & Schaub, 1993). Roughly speaking, such a logic is built by adding to first-order logic two distinct modalities, a "minimal belief" modality $B$ and a "negation as failure" modality *not*. The logic thus obtained is characterized in terms of a nice model-theoretic semantics. MBNF has been used in order to give a declarative semantics to very general classes of logic programs (Lifschitz & Woo, 1992; Schwarz & Lifschitz, 1993; Inoue & Sakama, 1994), which generalize the stable model semantics of negation as failure in logic programming (Gelfond & Lifschitz, 1988, 1990, 1991). Also, MBNF can be viewed as an extension of the theory of epistemic queries to databases (Reiter, 1990), which deals with the problem of querying a first-order database about its own knowledge. Due to its ability of expressing many features of nonmonotonic logics (Lifschitz, 1994; Schwarz & Lifschitz, 1993), MBNF is generally considered as a unifying framework for several nonmonotonic formalisms, including default logic, autoepistemic logic, circumscription, epistemic queries, and logic programming.

Although several aspects of the logic MBNF have been thoroughly investigated (Schwarz & Lifschitz, 1993; Chen, 1994; Bochman, 1995), the existing studies concerning the computational properties of MBNF are limited to subclasses of propositional MBNF theories (Inoue & Sakama, 1994) or to a very restricted subset of the first-order case (Beringer & Schaub, 1993).

In this paper we present a computational characterization of deduction in the propositional fragment of MBNF. In particular, we show that logical implication in the propositional fragment of MBNF is a $\Pi_3^p$-complete problem: hence, it is harder (unless the polynomial hierarchy collapses) than reasoning in all the best known propositional formalisms for nonmonotonic reasoning, like autoepistemic logic (Niemelä, 1992; Gottlob, 1992), default logic (Gottlob, 1992), circumscription (Eiter & Gottlob, 1993), (disjunctive) logic programming (Eiter & Gottlob, 1995), and several McDermott and Doyle's logics (Marek & Truszczyński, 1993). As shown in the following, this result also implies that minimal knowledge is computationally harder than negation as failure.

Moreover, we study the subclass of *flat* MBNF theories, i.e. MBNF theories without nested occurrences of modalities, showing that in this case logical implication is $\Pi_2^p$-complete. This case is the most interesting one from the logic programming viewpoint. Indeed, it implies that, under the stable model semantics, increasing the syntax of program rules, by allowing propositional formulas as goals in the rules, does not affect the worst-case complexity of query answering for disjunctive logic programs with negation as failure.

Furthermore, we provide algorithms for reasoning both in MBNF and in its flat fragment, which are optimal with respect to worst-case complexity. Notably, such deductive methods can be considered as generalizations of known methods for reasoning in nonmonotonic formalisms such as default logic, autoepistemic logic, and logic programming under stable model semantics.

We also show that the "negation as failure" modality in MBNF exactly corresponds to negative introspection in autoepistemic logic (Moore, 1985). This result implies that the logic MBNF can be considered as the "composition" of two epistemic modalities: the "minimal knowledge" operator due to Halpern and Moses (1985) and Moore's autoepistemic operator.

Besides its theoretical interest, we believe that such a computational and epistemological analysis of MBNF has interesting implications for the development of knowledge representation systems with nonmonotonic abilities, since it allows for a better understanding and comparison of the different nonmonotonic formalisms captured by MBNF. The interest in defining deductive methods for MBNF also arises from the fact that such a logic, originally developed as a framework for the comparison of different logical approaches to nonmonotonic reasoning, has recently been considered as an attractive knowledge representation formalism. In particular, it has been shown (Donini, Nardi, & Rosati, 1997a) that the full power of MBNF is necessary in order to logically formalize several features of implemented frame-based knowledge representation systems.

In the following, we first briefly recall the logic MBNF. In Section 3 we address the relationship between MBNF and Moore's autoepistemic logic. Then, in Section 4 we study the problem of reasoning in propositional MBNF: we first consider the case of general MBNF theories, then we deal with flat MBNF theories. In Section 5 we present the computational characterization of reasoning in MBNF. We conclude in Section 6. This paper is an extended and thoroughly revised version of (Rosati, 1997).

## 2. The Logic MBNF

In this section we briefly recall the logic MBNF (Lifschitz, 1994), which is a modal logic with two epistemic operators: a "minimal belief" modality $B$ and a "negation as failure" (also called "negation by default") modality $not$. We use $\mathcal{L}$ to denote a fixed propositional language built in the usual way from: (i) an alphabet $\mathcal{A}$ of propositional symbols; (ii) the symbols true, false; (iii) the propositional connectives $\vee, \wedge, \neg, \supset$. We denote as $\mathcal{L}_M$ the modal extension of $\mathcal{L}$ with the modalities $B$ and $not$. We say that a formula $\varphi \in \mathcal{L}_M$ has *(modal) depth $i$* (with $i \geq 0$) if each subformula in $\varphi$ lies within the scope of at most $i$ modalities, and there exists a subformula in $\varphi$ which lies within the scope of exactly $i$ modalities.

We denote as $\mathcal{L}_M^S$ the set of *subjective* MBNF formulas, i.e. the subset of formulas from $\mathcal{L}_M$ in which each occurrence of a propositional symbol lies within the scope of at least one modality, and with $\mathcal{L}_M^1$ the set of *flat* MBNF formulas, that is the set of formulas from $\mathcal{L}_M$ in which each propositional symbol lies within the scope of exactly one modality. We call a modal formula $\varphi$ from $\mathcal{L}_M$ *positive* (resp. *negative*) if the modality $not$ (resp. $B$) does not occur in $\varphi$. $\mathcal{L}_B$ denotes the set of positive formulas from $\mathcal{L}_M$, while $\mathcal{L}_B^S$ denotes the set of positive formulas from $\mathcal{L}_M^S$.

We now recall the notion of MBNF model. An *interpretation* is a set of propositional symbols. An MBNF *structure* is a triple $(I, M_b, M_n)$, where $I$ is an interpretation (also called initial world) and $M_b, M_n$ are non-empty sets of interpretations (worlds). Satisfiability of a formula in an MBNF structure is defined inductively as follows:

1. if $\varphi$ is a propositional symbol, $\varphi$ is satisfied by $(I, M_b, M_n)$ iff $\varphi \in I$;

2. $\neg \varphi$ is satisfied by $(I, M_b, M_n)$ iff $\varphi$ is not satisfied by $(I, M_b, M_n)$;

3. $\varphi_1 \wedge \varphi_2$ is satisfied by $(I, M_b, M_n)$ iff $\varphi_1$ is satisfied by $(I, M_b, M_n)$ and $\varphi_2$ is satisfied by $(I, M_b, M_n)$;

4. $\varphi_1 \vee \varphi_2$ is satisfied by $(I, M_b, M_n)$ iff either $\varphi_1$ is satisfied by $(I, M_b, M_n)$ or $\varphi_2$ is satisfied by $(I, M_b, M_n)$;

5. $\varphi_1 \supset \varphi_2$ is satisfied by $(I, M_b, M_n)$ iff either $\varphi_1$ is not satisfied by $(I, M_b, M_n)$ or $\varphi_2$ is satisfied by $(I, M_b, M_n)$;

6. $B\varphi$ is satisfied by $(I, M_b, M_n)$ iff, for every $J \in M_b$, $\varphi$ is satisfied by $(J, M_b, M_n)$;

7. $not\,\varphi$ is satisfied by $(I, M_b, M_n)$ iff there exists $J \in M_n$ such that $\varphi$ is not satisfied by $(J, M_b, M_n)$.

We write $(I, M_b, M_n) \models \varphi$ to indicate that $\varphi$ is satisfied by $(I, M_b, M_n)$. We say that a theory $\Sigma \subseteq \mathcal{L}_M$ is satisfied by $(I, M_b, M_n)$ (and write $(I, M_b, M_n) \models \Sigma$) iff each formula from $\Sigma$ is satisfied by $(I, M_b, M_n)$. If $\varphi \in \mathcal{L}_M^S$, then the evaluation of $\varphi$ is insensitive to the initial interpretation $I$: thus, in this case we also write $(M_b, M_n) \models \varphi$. Analogously, if $\varphi \in \mathcal{L}_B^S$, then the evaluation of $\varphi$ is insensitive both to the initial interpretation $I$ and to the set $M_n$, and we also write $M_b \models \varphi$. If $\varphi \in \mathcal{L}$ then the evaluation of $\varphi$ does not depend on the sets $M_b, M_n$, and in this case we write $I \models \varphi$.

In order to relate MBNF structures to standard interpretation structures in modal logic (i.e. Kripke structures), we remark that, due to the above notion of satisfiability, we can consider the sets $M_b$, $M_n$ in an MBNF interpretation structure as two distinct *universal* Kripke structures, i.e. possible-world structures in which each world is connected to all worlds of the structure. In fact, since the accessibility relation in such a structure is universal, without loss of generality it is possible to identify a universal Kripke structure with the set of interpretations contained in it. We recall that the class of universal Kripke structures characterizes the logic $\mathsf{S5}$ (Marek & Truszczyński, 1993, Theorem 7.52).

The nonmonotonic character of MBNF is obtained by imposing the following preference semantics over the interpretation structures satisfying a given theory.

**Definition 2.1** *A structure* $(I, M, M)$, *where* $M \neq \emptyset$, *is an* MBNF model *of a theory* $\Sigma \subseteq \mathcal{L}_M$ *iff* $(I, M, M) \models \Sigma$ *and, for each interpretation* $J$ *and for each set of interpretations* $M'$, *if* $M' \supset M$ *then* $(J, M', M) \not\models \Sigma$.

We say that a formula $\varphi$ is *entailed* (or *logically implied*) by $\Sigma$ in MBNF (and write $\Sigma \models_{\text{MBNF}} \varphi$) iff $\varphi$ is satisfied by every MBNF model of $\Sigma$. In order to simplify notation, we denote the MBNF model $(I, M, M)$ with the pair $(I, M)$, and, if $\Sigma \in \mathcal{L}_M^S$, we denote $(I, M, M)$ with $M$, since in this case the evaluation of $\Sigma$ is insensitive to the initial world $I$, namely, if $(I, M)$ is a model for $\Sigma$, then, for each interpretation $J$, $(J, M)$ is a model for $\Sigma$.

**Example 2.2** Let $\Sigma = \{Bp\}$. The only MBNF models for $\Sigma$ are of the form $(I, M)$, with $M = \{I : I \models p\}$. Hence, $\Sigma \models_{\text{MBNF}} Bp$, and $\Sigma \models_{\text{MBNF}} \neg B\psi$ for each $\psi \in \mathcal{L}$ such that the propositional formula $p \supset \psi$ is not valid. Therefore, the agent modeled by $\Sigma$ has minimal belief, in the sense that she only believes $p$ and the objective facts logically implied by $p$. □

**Example 2.3** Let $\Sigma = \{not\,married \supset B\,hasNoChildren\}$. It is easy to see that the only models for $\Sigma$ are of the form $(I, M)$ such that $M = \{I : I \models hasNoChildren\}$, since *married* can be assumed not to hold by the agent modeled by $\Sigma$, which is then able to conclude $B\,hasNoChildren$. Notably, the meaning of $\Sigma$ is analogous to the default rule $\frac{:\neg married}{hasNoChildren}$ in Reiter's default logic (Lifschitz, 1994). Also, let $\Sigma = \{B\,bird \wedge not\neg\,flies \supset B\,flies, B\,bird\}$. In a way analogous to the previous case, it can be shown that the only MBNF models for $\Sigma$ are of the form $(I, M)$, with $M = \{I : I \models bird \wedge flies\}$. Therefore, $\Sigma \models_{\text{MBNF}} B\,flies$. As shown by Lifschitz (1994), $\Sigma$ corresponds to the default theory $(\{\frac{bird:flies}{flies}\}, bird)$. □

Given a set of interpretations $M$, $Th(M)$ denotes the set of formulas $B\varphi$ such that $B\varphi \in \mathcal{L}_B$ and $M \models B\varphi$. Let $M_1, M_2$ be sets of interpretations. We say that $M_1$ is *equivalent* to $M_2$ iff $Th(M_1) = Th(M_2)$.

**Definition 2.4** *A set of interpretations* $M$ *is* maximal *iff, for each set of interpretations* $M'$, *if* $M'$ *is equivalent to* $M$ *then* $M' \subseteq M$.

It turns out that, when restricting to theories composed of subjective positive formulas, MBNF corresponds to the modal logic of minimal knowledge due to Halpern and Moses (1985), also known as ground nonmonotonic modal logic $\mathsf{S5}_G$ (Kaminski, 1991; Donini, Nardi, & Rosati, 1997b). In fact, $\mathsf{S5}_G$ is obtained from modal logic $\mathsf{S5}$ by imposing the following preference order over the universal Kripke structures satisfying a theory $\Sigma \in \mathcal{L}_B$: $M$ is a model for $\Sigma$ iff $M \models \Sigma$ and, for each $M'$, if $M' \models \Sigma$ then $M' \not\supseteq M$ (Shoham, 1987). In fact, it is immediate to see that the MBNF semantics of theories composed of subjective positive formulas corresponds to the above semantics according to $\mathsf{S5}_G$. Hence, the following property holds.

**Proposition 2.5** *Let* $\Sigma \subseteq \mathcal{L}_B$. *Then,* $M$ *is an* $\mathsf{S5}_G$ *model for* $\Sigma$ *iff, for each* $I$, $(I, M)$ *is an MBNF model for* $\{B\varphi : \varphi \in \Sigma\}$.

The previous proposition implies that, when $\Sigma \subseteq \mathcal{L}_B^S$, a set of interpretations $M$ satisfying $\Sigma$ is compared with all other sets of interpretations satisfying $\Sigma$, while, in the case $\Sigma \subseteq \mathcal{L}_M$, $M$ is only compared with the sets $M'$ such that $(M', M)$ satisfies $\Sigma$.

Hence, the main difference between MBNF and $\mathsf{S5}_G$ lies in the fact that in $\mathsf{S5}_G$ all models are maximal with respect to set containment (or minimal with respect to the objective knowledge which holds in the model), while in MBNF this property is not generally true. E.g., the theory $\Sigma = \{not\ married \lor B\ married\}$ has two types of models, for each possible choice of the initial world $J$: $(J, M_1)$, where $M_1$ corresponds to the set of all interpretations (which represents the case in which $married$ is not assumed to hold); and $(J, M_2)$, where $M_2 = \{I : I \models married\}$. Namely, if $married$ is assumed to hold, then $\Sigma$ forces one to conclude $B\ married$: that is, the initial assumption is *justified* by the knowledge derived on the basis of such an assumption (Lin & Shoham, 1992). We remark that, by Proposition 2.5, the interpretation of the MBNF operator $B$ exactly corresponds to the interpretation of the modality $B$ in $\mathsf{S5}_G$.

## 3. Relating MBNF to Autoepistemic Logic

In this section we study the relationship between autoepistemic logic and MBNF. First, we briefly recall Moore's autoepistemic logic (AEL). In order to keep notation to a minimum, we change the language of AEL, using the modality $B$ instead of $L$. Thus, in the following a formula of AEL is a formula from $\mathcal{L}_B$.

**Definition 3.1** *A propositionally consistent set of formulas* $T \subset \mathcal{L}_B$ *is a stable expansion for a set of initial knowledge* $\Sigma \subseteq \mathcal{L}_B$ *if* $T$ *satisfies the following equation:*

$$T = Cn(\Sigma \cup \{B\varphi \mid \varphi \in T\} \cup \{\neg B\varphi \mid \varphi \notin T\}) \tag{1}$$

*where* $Cn(S)$ *denotes the propositional deductive closure of the modal theory* $S \subseteq \mathcal{L}_B$.

Given a theory $\Sigma \subseteq \mathcal{L}_B$ and a formula $\varphi \in \mathcal{L}_B$, we write $\Sigma \models_{\mathrm{AEL}} \varphi$ iff $\varphi$ belongs to all the stable expansions of $\Sigma$. Each stable expansion $T$ is a *stable set* according to the following definition (Stalnaker, 1993).

**Definition 3.2** *A modal theory* $T \subseteq \mathcal{L}_B$ *is a stable set if*

1. $T = Cn(T)$;

2. for every $\varphi \in \mathcal{L}_B$, if $\varphi \in T$ then $B\varphi \in T$;

3. for every $\varphi \in \mathcal{L}_B$, if $\varphi \notin T$ then $\neg B\varphi \in T$.

We recall that a stable set $T$ corresponds to a maximal universal Kripke structure $M_T$ such that $T$ is the set of formulas satisfied by $M_T$ (Marek & Truszczyński, 1993). With the term AEL *model* for $\Sigma$ we will thus refer to a set of interpretations $M$ whose set of theorems $Th(M)$ corresponds to a stable expansion for $\Sigma$ in AEL.

Finally, notice that we have adopted the notion of *consistent* autoepistemic logic (Marek & Truszczyński, 1993), i.e. in (1) we do not allow the inconsistent theory $T = \mathcal{L}_B$ composed of all modal formulas to be a (possible) stable expansion. The results presented in this section can be easily extended to this case (corresponding to Moore's original proposal): however, this requires to slightly change the semantics of MBNF, allowing in Definition 2.1 the empty set of interpretations to be a possible component of MBNF structures.

In the following, we use the term *embedding* (or translation) to indicate a transformation function $\tau(\cdot)$ for modal theories. We are interested in finding a *faithful* embedding (Gottlob, 1995; Schwarz, 1996; Janhunen, 1998), in the following sense: $\tau(\cdot)$ is a faithful embedding of AEL into MBNF if, for each theory $\Sigma \subseteq \mathcal{L}_B$ and for each model $M$, $M$ is an AEL model for $\Sigma$ iff $M$ is an MBNF model for $\tau(\Sigma)$.

It is already known that AEL theories can be embedded into MBNF theories. In particular, it has been proven (Lin & Shoham, 1992; Schwarz & Truszczyński, 1994) that AEL theories with no nested occurrences of $B$ (called *flat* theories) can be embedded into MBNF; now, since in AEL any theory can be transformed into an equivalent flat theory (which has in general size exponential in the size of the initial theory), it follows that any AEL theory can be embedded into MBNF.

However, we now prove a much stronger result: negation as failure in MBNF *exactly* corresponds to negative introspection in AEL, i.e. AEL's modality $\neg B$ and MBNF's modality *not* are semantically equivalent. Hence, such a correspondence is not only limited to modal theories without nested modalities, and induces a polynomial-time embedding of *any* AEL theory into MBNF.

We first define the translation $\tau(\cdot)$ of modal theories from AEL to MBNF theories.

**Definition 3.3** Let $\varphi \in \mathcal{L}_B$. Then, $\tau(\varphi)$ is the MBNF formula obtained from $\varphi$ by substituting each occurrence of $B$ with $\neg not$. Moreover, if $\Sigma \subseteq \mathcal{L}_B$, then $\tau(\Sigma)$ denotes the MBNF theory $\{B\tau(\varphi)|\varphi \in \Sigma\}$.

We now show that the translation $\tau(\cdot)$ embeds AEL theories into MBNF. To this aim, we exploit the semantic characterization of AEL defined by Schwarz (1992). Roughly speaking, according to such a preference semantics over possible-world structures, an AEL model for $\Sigma$ is a set of interpretations $M$ satisfying $\Sigma$ such that, for any interpretation $J$ not contained in $M$, the pair $(J, M)$ does not satisfy $\Sigma$. Formally:

**Proposition 3.4** (Schwarz, 1992, Proposition 4.1) Let $\Sigma \subseteq \mathcal{L}_B$. Then, $M$ is an AEL model for $\Sigma$ iff, for each interpretation $I \in M$, $(I, M) \models \Sigma$ and, for each interpretation $J \notin M$, $(J, M) \not\models \Sigma$.

In the following, we say that an occurrence of a subformula $\psi$ in a formula $\varphi \in \mathcal{L}_M$ is *strict* if it does not lie within the scope of a modal operator. E.g., let $\sigma = B\varphi \wedge not(B\psi \vee \xi)$. The occurrence of $B\varphi$ in $\sigma$ is strict, while the occurrence of $B\psi$ is not strict.

**Theorem 3.5** *Let* $\Sigma \subseteq \mathcal{L}_B$. *Then,* $M$ *is an* AEL *model for* $\Sigma$ *iff, for each* $I$, $(I, M)$ *is an* MBNF *model of* $\tau(\Sigma)$.

*Proof.* *If part.* Suppose $(I, M)$ is an MBNF model of $\tau(\Sigma)$. Then, for each $M' \supset M$, $(M', M) \not\models \tau(\Sigma)$. Since $\tau(\Sigma)$ is a set of formulas of the form $B\varphi$, where $\varphi$ does not contain any occurrence of the operator $B$, it follows that, for each subformula of the form $not\, \varphi$ occurring in $\tau(\Sigma)$, $(M', M) \models not\, \varphi$ iff $(M, M) \models not\, \varphi$. Now let $B\varphi \in \tau(\Sigma)$, let $\varphi'$ denote the propositional formula obtained from $\varphi$ by replacing each strict occurrence in $\varphi$ of a formula of the form $not\, \psi$ with true if $(M, M) \models not\, \psi$ and with false otherwise, and let $\Sigma' = \{\varphi' : B\varphi \in \tau(\Sigma)\}$. Now suppose there exists an interpretation $J$ such that $J \models \Sigma'$ and $J \notin M$. Then, from the definition of satisfiability in MBNF structures it follows that $(M \cup \{J\}, M) \models \tau(\Sigma)$, thus contradicting the hypothesis that $(I, M)$ is an MBNF model for $\tau(\Sigma)$. Hence, $M = \{I : I \models \Sigma'\}$. Now consider a pair $(J, M)$: again, from the definition of satisfiability in MBNF structures it follows immediately that $(J, M) \models \Sigma$ iff $J \models \Sigma'$. And since $M$ contains all the interpretations satisfying $\Sigma'$, it follows that, for each interpretation $J \notin M$, $(J, M) \not\models \Sigma$, therefore by Proposition 3.4 it follows that $M$ is an AEL model for $\Sigma$.

*Only-if part.* Suppose $M$ is an AEL model for $\Sigma$. Then, by Proposition 3.4, for each interpretation $I \in M$, $(I, M) \models \Sigma$ and, for each interpretation $J \notin M$, $(J, M) \not\models \Sigma$. For each $\varphi \in \Sigma$, let $\varphi''$ denote the propositional formula obtained from $\varphi$ by replacing each strict occurrence of a formula of the form $B\psi$ with true if $M \models B\psi$ and with false otherwise, and let $\Sigma'' = \{\varphi'' : \varphi \in \Sigma\}$. Then, suppose there exists an interpretation $J$ such that $J \models \Sigma''$ and $J \notin M$. Then, from the definition of satisfiability in MBNF structures it follows that $(J, M) \models \Sigma$, thus contradicting the hypothesis that $M$ is an AEL model for $\Sigma$. Hence, $M = \{I : I \models \Sigma''\}$. Now suppose that, for some interpretation $I$, $(I, M)$ is not an MBNF model for $\tau(\Sigma)$. Then, there exists $M' \supset M$ such that $(M', M) \models \tau(\Sigma)$. From the definition of $\tau(\cdot)$, it follows that each interpretation in $M'$ satisfies $\Sigma''$, and, since $M' \supset M$, there exists $J \notin M$ such that $J \models \Sigma''$. Contradiction. Therefore, $(I, M)$ is an MBNF model for $\Sigma$. $\qquad\square$

We remark that the above theorem could alternatively be proved from the fact that the $K$-free fragment of the logic MKNF (Lifschitz, 1991) is equivalent to AEL, which is stated (although without proof) by Schwarz and Truszczyński (1994, page 123), and from the correspondence between MBNF and MKNF (Lifschitz, 1994).

The previous theorem implies that the interpretation of the modality $not$ in MBNF and of the modal operator in autoepistemic logic are the same. This property extends previous results relating MBNF with AEL (Lin & Shoham, 1992; Schwarz & Lifschitz, 1993; Chen, 1994), and has interesting consequences both in the logic programming framework and in nonmonotonic reasoning. In particular, since MBNF generalizes the stable model semantics for logic programs (Gelfond & Lifschitz, 1988), the above result strengthens the idea that AEL is the true logic of negation as failure (as interpreted according to the stable model semantics). Moreover, positive theories have the same interpretation both in MBNF and in the logic of minimal knowledge $\mathsf{S5}_G$ (Halpern & Moses, 1985): consequently, the logic MBNF generalizes both Halpern and Moses' $\mathsf{S5}_G$ and Moore's AEL.

## 4. Reasoning in MBNF

In this section we present algorithms for reasoning in propositional MBNF: in particular, we study the entailment problem in MBNF. From now on, we assume to deal with *finite* MBNF theories $\Sigma$, therefore we refer to a single formula $\sigma$ (which corresponds to the conjunction of all the formulas contained in the finite theory $\Sigma$).

### 4.1 Characterizing MBNF Models

We now present a finite characterization of the MBNF models of a formula $\sigma \in \mathcal{L}_M$. As in several methods for reasoning in nonmonotonic modal logics (Gottlob, 1992; Marek & Truszczyński, 1993; Eiter & Gottlob, 1992; Niemelä, 1992; Donini et al., 1997b), the technique we employ is based on the definition of a correspondence between the preferred models of a theory and the partitions of the set of modal subformulas of the theory. In fact, such partitions can be used in order to provide a finite characterization of a universal Kripke structure: specifically, a partition satisfying certain properties identifies a particular universal Kripke structure $M$, by uniquely determining a propositional theory such that $M$ is the set of all interpretations satisfying such a theory.

We extend such known techniques in order to deal with the preference semantics of MBNF. In particular, we characterize the properties that a partition of modal subformulas of a formula $\sigma \in \mathcal{L}_M$ must satisfy in order to identify an MBNF model for $\sigma$. In this way, we provide a method that does not rely on a modal logic theorem prover, but reduces the problem of reasoning in a bimodal logic to a number of reasoning problems in propositional logic.

First, we introduce some preliminary definitions. We call a formula of the form $B\varphi$ or $not \, \varphi$, with $\varphi \in \mathcal{L}_M$, a *modal atom*.

**Definition 4.1** *Let $\sigma \in \mathcal{L}_M$. We call the set of modal atoms occurring in $\sigma$ the* modal atoms of $\sigma$ *(and denote such a set as $MA(\sigma)$).*

**Definition 4.2** *Let $\sigma \in \mathcal{L}_M$ and let $(P, N)$ be a partition of a set of modal atoms. We denote as $\sigma(P, N)$ the formula obtained from $\sigma$ by substituting each strict occurrence in $\sigma$ of a formula in $P$ with* true, *and each strict occurrence in $\sigma$ of a formula in $N$ with* false.

Observe that only the occurrences in $\sigma$ of modal subformulas which are not within the scope of another modality are replaced; notice also that, if $P \cup N$ contains $MA(\sigma)$, then $\sigma(P, N)$ is a propositional formula. In this case, the pair $(P, N)$ identifies a guess on the modal subformulas from $\sigma$, i.e. $P$ contains the modal subformulas of $\sigma$ assumed to hold, while $N$ contains the modal subformulas of $\sigma$ assumed not to hold.

**Definition 4.3** *Let $\sigma \in \mathcal{L}_M$ and let $(P, N)$ be a partition of $MA(\sigma)$. We denote as $ob(P, N)$ the propositional formula*

$$ob(P, N) = \bigwedge_{B\varphi \in P} \varphi(P, N)$$

Roughly speaking, the propositional formula $ob(P, N)$ represents the "objective knowledge" implied by the guess $(P, N)$ on the formulas of the form $B\varphi$ belonging to $P$. From the semantic viewpoint, in each structure $(I, M, M')$ satisfying the guess on the modal atoms given by $(P, N)$, the propositional formula $ob(P, N)$ constrains the interpretations of $M$, since in each such structure the propositional formula $ob(P, N)$ must be satisfied by each interpretation $J \in M$, i.e. $J \models ob(P, N)$.

**Example 4.4** Let

$$\sigma = (Ba \lor not(b \land c)) \land d \land \neg B(\neg f \lor g)$$

Then, $MA(\sigma) = \{Ba, not(b \land c), B(\neg f \lor g)\}$. Now suppose that

$$
\begin{aligned}
P &= \{Ba, not(b \land c)\} \\
N &= \{B(\neg f \lor g)\}
\end{aligned}
$$

Then, $\sigma(P, N) = (\text{true} \lor \text{true}) \land d \land \neg\text{false}$ (which is equivalent to $d$), and $ob(P, N) = a$. $\square$

**Definition 4.5** *We say that a pair of sets of interpretations $(M, M')$ induces the partition $(P, N)$ of $MA(\sigma)$ if, for each modal atom $\xi \in MA(\sigma)$, $\xi \in P$ iff $(M, M') \models \xi$.*

**Lemma 4.6** *Let $\varphi \in \mathcal{L}_M$, let $I$ be an interpretation, let $M, M'$ be sets of interpretations, and let $(P, N)$ be the partition induced by $(M, M')$ on a set of modal atoms $S$. Then, $(I, M, M') \models \varphi$ iff $(I, M, M') \models \varphi(P, N)$.*

*Proof.* Follows immediately from Definitions 4.2 and 4.5, and from the definition of satisfiability in MBNF structures. $\square$

We now show that, if $(I, M)$ is an MBNF model for $\sigma$ which induces the partition $(P, N)$ of $MA(\Sigma)$, then the formula $ob(P, N)$ completely characterizes the set of interpretations $M$.

**Theorem 4.7** *Let $\sigma \in \mathcal{L}_M$, let $(I, M)$ be an MBNF model for $\sigma$, and let $(P, N)$ be the partition of $MA(\sigma)$ induced by $(M, M)$. Then, $M = \{J : J \models ob(P, N)\}$.*

*Proof.* Let $M' = \{J : J \models ob(P, N)\}$. Since $(M, M)$ induces the partition $(P, N)$, by Definition 4.5 it follows that each interpretation in $M$ must satisfy $ob(P, N)$, hence $M \subseteq M'$. Now suppose $M \subset M'$, and consider the structure $(I, M', M)$. We prove that each modal atom $\xi \in MA(\sigma)$ belongs to $P$ iff $(I, M', M) \models \xi$. The proof is by induction on the depth of formulas in $MA(\sigma)$.

First, consider a modal atom $not\,\psi$ such that $\psi \in \mathcal{L}$: from the definition of satisfiability of a formula in an MBNF structure, it follows immediately that $not\,\psi \in P$ iff $(I, M', M) \models not\,\psi$. Then, consider a modal atom $B\psi$ such that $\psi \in \mathcal{L}$: if $B\psi \in P$, then, by definition of $ob(P, N)$, the propositional formula $ob(P, N) \supset \psi$ is valid, therefore $(I, M', M) \models B\psi$. If $B\psi \in N$, then there exists an interpretation $J$ in $M$ such that $J \not\models \psi$, and since $M' \supset M$, it follows that $(I, M', M) \not\models B\psi$. Hence, each modal atom $\xi \in MA(\sigma)$ of depth 1 belongs to $P$ iff $(I, M', M) \models \xi$.

Suppose now that $\xi \in P$ iff $(I, M', M) \models \xi$ for each modal atom $\xi$ in $MA(\sigma)$ of depth less or equal to $i$. Consider a modal atom $B\psi$ of $MA(\sigma)$ of depth $i + 1$: by the induction hypothesis, and by Lemma 4.6, $(I, M', M) \models B\psi$ iff $M' \models B(\psi(P, N))$. Now, if $B\psi \in P$, then, by definition of $ob(P,N)$, the propositional formula $ob(P,N) \supset \psi(P,N)$ is valid, and since $M' = \{J : J \models ob(P,N)\}$, it follows that $M' \models B(\psi(P,N))$, which in turn implies $(I, M', M) \models B\psi$; on the other hand, if $B\psi \in N$, then there exists an interpretation $J$ in $M$ such that $(J, M, M) \not\models \psi$, hence, by the induction hypothesis and Lemma 4.6, $J \not\models \psi(P,N)$. Now, since $M' \supset M$, it follows that $M' \not\models B(\psi(P,N))$, hence $(I, M', M) \not\models B\psi$. In the same way it is possible to show that a modal atom of the form $not\ \psi$ of depth $i+1$ belongs to $P$ iff $(I, M', M) \models not\ \psi$.

We have thus proved that each modal atom $\xi \in MA(\sigma)$ belongs to $P$ iff $(I, M', M) \models \xi$: this in turn implies that $(I, M', M) \models \sigma$ iff $I \models \sigma(P,N)$, and since by hypothesis $(I, M, M)$ satisfies $\sigma$ and $(P, N)$ is the partition of $MA(\Sigma)$ induced by $(M, M)$, by Lemma 4.6 it follows that $I \models \sigma(P,N)$. Therefore, $(I, M', M) \models \sigma$, which contradicts the hypothesis that $(I, M)$ is an MBNF model for $\sigma$. Consequently, $M' = M$, which proves the thesis. $\qquad\square$

Informally, the above theorem states that each MBNF model for $\sigma$ can be associated with a partition $(P, N)$ of the modal atoms of $\sigma$; moreover, the propositional formula $ob(P,N)$ exactly characterizes the set of interpretations $M$ of an MBNF model $(I, M)$, in the sense that $M$ is the set of *all* interpretations satisfying $ob(P,N)$. This provides a finite way to describe all MBNF models for $\sigma$.

We now define the notion of a partition of a set of modal atoms induced by a pair of propositional formulas.

**Definition 4.8** *Let $\sigma \in \mathcal{L}_M$, $\varphi_1, \varphi_2 \in \mathcal{L}$. We denote as $Prt(\sigma, \varphi_1, \varphi_2)$ the partition of $MA(\sigma)$ induced by $(M_1, M_2)$, where $M_1 = \{I : I \models \varphi_1\}$, $M_2 = \{I : I \models \varphi_2\}$.*

In order to simplify notation, we denote as $Prt(\sigma, \varphi)$ the partition $Prt(\sigma, \varphi, \varphi)$. The following theorem provides a constructive way to build the partition $Prt(\sigma, \varphi, \psi)$.

**Theorem 4.9** *Let $\sigma \in \mathcal{L}_M$, $\varphi, \psi \in \mathcal{L}$. Let $(P, N)$ be the partition of $MA(\sigma)$ built as follows:*

1. *start from $P = N = \emptyset$;*

2. *for each modal atom $B\xi$ in $MA(\sigma)$ such that $\xi(P,N) \in \mathcal{L}$, if the propositional formula $\varphi \supset \xi(P,N)$ is valid, then add $B\xi$ to $P$, otherwise add $B\xi$ to $N$;*

3. *for each modal atom $not\ \xi$ in $MA(\sigma)$ such that $\xi(P,N) \in \mathcal{L}$, if the propositional formula $\psi \supset \xi(P,N)$ is not valid, then add $not\ \xi$ to $P$, otherwise add $not\ \xi$ to $N$;*

4. *iteratively apply the above rules until $P \cup N = MA(\sigma)$.*

*Then, $(P, N) = Prt(\sigma, \varphi, \psi)$.*

**Proof.** The proof is by induction on the structure of the formulas in $MA(\sigma)$. First, from the fact that $Prt(\sigma, \varphi, \psi)$ is the partition induced by $(M, M')$, with $M = \{I : I \models \varphi\}$, $M' = \{I : I \models \psi\}$, and from the definition of satisfiability in MBNF structures, it follows that, if $\xi \in \mathcal{L}$, then $(M, M') \models B\xi$ if and only if $\varphi \supset \xi$ is a valid propositional formula, and $(M, M') \models not\ \xi$ if and only if $\psi \supset \xi$ is not a valid propositional formula. Therefore, $(P, N)$ agrees with $Prt(\sigma, \varphi, \psi)$ on all modal atoms of modal depth 1. Suppose now that $(P, N)$ and $Prt(\sigma, \varphi, \psi)$ agree on all modal atoms of modal depth less or equal to $i$. Consider a modal atom $B\xi$ of $MA(\sigma)$ of modal depth $i+1$. From Lemma 4.6 and from the definition of satisfiability in MBNF structures, it follows that $(M, M') \models B\xi$ if and only if $\varphi \supset \xi(Prt(\sigma, \varphi, \psi))$ is a valid propositional formula, and since by Definition 4.2 the value of the formula $\xi(Prt(\sigma, \varphi, \psi))$ only depends on the guess of the modal atoms of modal depth less or equal to $i$ in $Prt(\sigma, \varphi, \psi)$, by the induction hypothesis it follows that $\xi(Prt(\sigma, \varphi, \psi)) = \xi(P, N)$, hence $B\xi$ belongs to $P$ if and only if $(M, M') \models B\xi$. Analogously, it can be proven that any modal atom of depth $i+1$ of the form $not\ \xi$ belongs to $P$ if and only if $(M, M') \models not\ \xi$. Therefore, $(P, N)$ and $Prt(\sigma, \varphi, \psi)$ agree on all modal atoms of modal depth $i+1$. $\square$

The algorithms we present in the following for reasoning in MBNF use the above shown properties of partitions of modal subformulas of a formula $\sigma$, together with additional conditions on such partitions (that vary according to the different classes of theories accepted as inputs), in order to identify all the MBNF models for $\sigma$.

As for the entailment problem $\sigma \models_{\mathrm{MBNF}} \varphi$, we point out that the occurrences of $not$ in $\varphi$ are equivalent to occurrences of $\neg B$, since in each MBNF model for $\sigma$ both modalities in $\varphi$ are evaluated on the same set of interpretations. Therefore, as in the original formulation of MBNF (Lifschitz, 1994), we restrict query answering in MBNF to positive formulas.

Let $\varphi \in \mathcal{L}_B$, $\psi \in \mathcal{L}$, and $M = \{J : J \models \psi\}$. We denote as $\varphi(\psi)$ the propositional formula obtained from $\varphi$ by substituting each strict occurrence of a modal atom $B\xi$ of $\varphi$ with true if $M \models B\xi$, and with false otherwise. It can be immediately verified that $\varphi(\psi) = \varphi(Prt(\varphi, \psi))$.

**Theorem 4.10** *Let $\sigma, \varphi \in \mathcal{L}_M$. Let $(I, M)$ be an MBNF model for $\sigma$ and let $(P, N)$ be the partition of $MA(\sigma)$ induced by $(M, M)$. Then, $\varphi$ is satisfied by $(I, M, M)$ iff $I \models \varphi(ob(P, N))$.*

*Proof.* The proof follows immediately from the fact that, by Theorem 4.9, $\varphi(ob(P, N)) = \varphi(Prt(\varphi, ob(P, N)))$, and from Lemma 4.6. $\square$

We now show that the entailment problem in MBNF is related to the membership problem for stable sets (Gottlob, 1995), which in turn is related to the notion of (objective) kernel that has been used to characterize stable expansions of autoepistemic theories (Marek & Truszczyński, 1993).

**Definition 4.11** *Let $\psi \in \mathcal{L}$. We denote as $ST(\psi)$ the (unique) stable set $T \subseteq \mathcal{L}_B$ such that*

$$T \cap \mathcal{L} = \{\varphi \in \mathcal{L} | \psi \supset \varphi \text{ is valid}\}$$

**Theorem 4.12** *Let $\sigma \in \mathcal{L}_M$, $\varphi \in \mathcal{L}_B^S$. Then, $\sigma \models_{\mathrm{MBNF}} \varphi$ iff there exists an MBNF model $(I, M)$ for $\sigma$ such that $\varphi \notin ST(ob(P, N))$, where $(P, N)$ is the partition of $MA(\sigma)$ induced by $(M, M)$.*

*Proof.* Let $M = \{I : I \models ob(P, N)\}$: from the above definition and Definition 3.2, it follows immediately that $ST(ob(P, N)) = Th(M)$. Therefore, if $\varphi \in \mathcal{L}_B^S$ then $(I, M, M) \models \varphi$ iff $\varphi \in ST(ob(P, N))$. $\square$

---

**Algorithm** MBNF-Not-Entails$(\sigma, \varphi)$
**Input:** formula $\sigma \in \mathcal{L}_M$, formula $\varphi \in \mathcal{L}_B$;
**Output:** true if $\sigma \not\models_{\mathrm{MBNF}} \varphi$, false otherwise.
**begin**
**if there exists** partition $(P, N)$ of $MA(\sigma)$
**such that**
    (a) $(P, N) = Prt(\Sigma, ob(P, N))$ **and**
    (b) $\sigma(P, N) \wedge \neg\varphi(ob(P, N))$ is satisfiable **and**
    (c) **for each** partition $(P', N') \neq (P, N)$ of $MA(\sigma)$,
        (c1) $\sigma(P', N')$ is not satisfiable **or**
        (c2) $(P', N') \neq Prt(\Sigma, ob(P', N'), ob(P, N))$ **or**
        (c3) $ob(P, N) \wedge \neg ob(P', N')$ is satisfiable
**then return** true
**else return** false
**end**

---

Figure 1: Algorithm MBNF-Not-Entails.

## 4.2 Reasoning in Propositional MBNF

We now define a deductive method for reasoning in general propositional MBNF theories. Specifically, we present the algorithm MBNF-Not-Entails, reported in Figure 1, for computing entailment in MBNF.

The algorithm exploits the finite characterization of MBNF models given by Theorem 4.7 and an analogous finite characterization, in terms of partitions of $MA(\sigma)$, of all the models relevant for establishing whether a partition $(P, N)$ of $MA(\sigma)$ identifies an MBNF model.

The algorithm checks whether there exists a partition $(P, N)$ of $MA(\sigma)$ satisfying the three conditions (a), (b), (c). Intuitively, the partition cannot be self-contradictory (condition (a)): in particular, the condition $(P, N) = Prt(\Sigma, ob(P, N))$ establishes that the objective knowledge implied by the partition $(P, N)$ (that is, the formula $ob(P, N)$) identifies a set of interpretations $M = \{I : I \models ob(P, N)\}$ such that $(M, M)$ induces the same partition $(P, N)$ on $MA(\sigma)$. Moreover, the partition must be consistent with $\sigma$ and $\neg\varphi$ (condition (b)): such a condition implies that there exists an interpretation $I$ such that both $\sigma$ is satisfied in $(I, M, M)$ and $\varphi$ is not satisfied in the structure $(I, M, M)$. Finally, condition (c) corresponds to check whether such a structure $(I, M, M)$ identifies an MBNF model for $\sigma$ according to Definition 2.1, i.e. whether there is no pair $(J, M')$ such that $M' \supset M$ and $(J, M', M)$ satisfies $\sigma$. Again, the search of such a structure is performed by examining whether there exists a partition of $MA(\sigma)$, different from $(P, N)$, which does not satisfy any of the conditions (c1), (c2), (c3).

We illustrate the algorithm through the following simple example.

**Example 4.13** Suppose

$$\sigma = B(a \vee Bb) \wedge (not(c \vee \neg d) \vee B\neg not\, b) \wedge c$$

$$\varphi \;=\; \neg Bb \vee (\neg b \wedge B(a \wedge b))$$

Then, $MA(\sigma) = \{B(a \vee Bb), Bb, not(c \vee \neg d), B\neg not\,b, not\,b\}$. Now suppose that $(P, N) = (P_1, N_1)$, where

$$
\begin{aligned}
P_1 \;&=\; \{B(a \vee Bb), not(c \vee \neg d), not\,b\}\\
N_1 \;&=\; \{Bb, B\neg not\,b\}
\end{aligned}
$$

Then, $\sigma(P, N) = \mathsf{true} \wedge (\mathsf{true} \vee \mathsf{false}) \wedge c$ (which is equivalent to $c$), and $ob(P, N) = a \vee \mathsf{false}$ (which is equivalent to $a$). Now, let $M = \{I : I \models a\}$: it is easy to see that $(M, M)$ satisfies the modal atoms in $P$, while it does not satisfy the modal atoms in $N$, hence $(P, N) = Prt(\Sigma, ob(P, N))$, thus satisfying condition (a) of the algorithm. Then, since $a \supset a \wedge b$ is not a valid propositional formula, $M \not\models B(a \wedge b)$, hence $\neg\varphi(ob(P, N)) = \neg(\mathsf{true} \vee (\neg b \wedge \mathsf{false}))$, which is equivalent to $\mathsf{false}$. Therefore, $\sigma(P, N) \wedge \neg\varphi(ob(P, N))$ is not satisfiable, thus condition (b) does not hold.

Suppose now that $(P, N) = (P_2, N_2)$, where

$$
\begin{aligned}
P_2 \;&=\; \{B(a \vee Bb), not(c \vee \neg d), Bb, B\neg not\,b\}\\
N_2 \;&=\; \{not\,b\}
\end{aligned}
$$

Then, $\sigma(P, N) = \mathsf{true} \wedge (\mathsf{true} \vee \mathsf{true}) \wedge c$ (which is equivalent to $c$), and $ob(P, N) = (a \vee \mathsf{true}) \wedge b \wedge \mathsf{true}$, which is equivalent to $b$. Again, it is easy to see that $(P, N) = Prt(\Sigma, ob(P, N))$, thus satisfying condition (a) of the algorithm. Then, since $b \supset a \wedge b$ is not a valid propositional formula, $\neg\varphi(ob(P, N)) = \neg(\mathsf{false} \vee (\neg b \wedge \mathsf{false}))$, which is equivalent to $\mathsf{true}$. Hence, $\sigma(P, N) \wedge \neg\varphi(ob(P, N))$ is equivalent to $c$, which implies that condition (b) holds. Finally, it is easy to verify that either condition (c1) or condition (c2) holds for each partition of $MA(\sigma)$ different from $(P_2, N_2)$, with the exception of $(P_1, N_1)$. So let $(P', N') = (P_1, N_1)$: as shown before, $ob(P', N')$ is equivalent to $a$, hence $ob(P, N) \wedge \neg ob(P', N')$ is equivalent to $b \wedge \neg a$, therefore condition (c3) holds for $(P', N') = (P_1, N_1)$, which implies that condition (c) holds for $(P, N) = (P_2, N_2)$. Consequently, MBNF-Not-Entails$(\sigma, \varphi)$ returns $\mathsf{true}$. In fact, the partition $(P_2, N_2)$ identifies the set of MBNF models for $\sigma$ $(I, M)$ such that $I$ is an interpretation satisfying $c$ and $M = \{I : I \models b\}$. Each such model does not satisfy the query $\varphi$: indeed, it can immediately be verified that, for each interpretation $I$, $(I, M, M) \not\models \neg Bb \vee (\neg b \wedge B(a \wedge b))$, since $M \not\models B(a \wedge b)$ and $M \models Bb$. $\qquad\square$

To prove correctness of the algorithm MBNF-Not-Entails we need the following preliminary lemma.

**Lemma 4.14** *Let $\sigma \in \mathcal{L}_M$, and let $(P, N)$ be the partition of $MA(\Sigma)$ induced by $(M', M)$. Let $M'' = \{I : I \models ob(P, N)\}$. Then, $(P, N)$ is the partition induced by $(M'', M)$.*

*Proof.* The proof is by induction on the depth of the modal atoms of $MA(\Sigma)$. Let $not\,\psi \in MA(\Sigma)$ such that $\psi \in \mathcal{L}$: then, $(M', M) \models not\,\psi$ iff there exists an interpretation $I \in M$ such that $I \not\models \psi$, therefore $(M', M) \models not\,\psi$ iff $(M'', M) \models not\,\psi$. Now let $B\psi \in MA(\Sigma)$ such that $\psi \in \mathcal{L}$: by Definition 4.3, $(M', M) \models B\psi$ iff the propositional formula $ob(P, N) \supset \psi$ is valid, and since $M'' = \{I : I \models ob(P, N)\}$, it follows that $(M', M) \models B\psi$ iff $(M'', M) \models B\psi$.

Now suppose that, for each modal atom $\xi$ of depth $i$, $(M', M) \models \xi$ iff $(M'', M) \models \xi$, and let $(P', N')$ denote the partition of the modal atoms in $MA(\Sigma)$ of depth less or equal to $i$ induced by $(M', M)$. First, consider a modal atom $not\ \psi$ of depth $i + 1$. Then, by Lemma 4.6, $(M', M) \models not\ \psi$ iff $(M', M) \models not(\psi(P', N'))$ and, by the inductive hypothesis and Lemma 4.6, $(M'', M) \models not\ \psi$ iff $(M'', M) \models not(\psi(P', N'))$. Then, since $\psi$ has depth $i$, $\psi(P', N')$ is a propositional formula, hence $(M', M) \models not(\psi(P', N'))$ iff there exists an interpretation $I \in M$ such that $I \not\models \psi(P', N')$, which immediately implies that $(M', M) \models not\ \psi$ iff $(M'', M) \models not\ \psi$. Now consider a modal atom $B\psi$ of depth $i + 1$. Then, by Lemma 4.6, $(M', M) \models B\psi$ iff $(M', M) \models B(\psi(P', N'))$ and, by the inductive hypothesis and Lemma 4.6, $(M'', M) \models B\psi$ iff $(M'', M) \models B(\psi(P', N'))$. By Definition 4.3, $(M', M) \models B\psi$ iff the propositional formula $ob(P, N) \supset \psi(P', N')$ is valid, and since $M'' = \{I : I \models ob(P, N)\}$, it follows that $(M', M) \models B\psi$ iff $(M'', M) \models B\psi$, which proves the thesis. $\square$

We are now ready to prove correctness of the algorithm MBNF-Not-Entails.

**Theorem 4.15** *Let* $\sigma \in \mathcal{L}_M$, $\varphi \in \mathcal{L}_B$. *Then,* MBNF-*Not-Entails*$(\sigma, \varphi)$ *returns* true *iff* $\sigma \not\models_{\text{MBNF}} \varphi$.

*Proof. If part.* Suppose $\sigma \not\models_{\text{MBNF}} \varphi$. Then, there exists a pair $(I, M)$ such that $(I, M)$ is an MBNF model for $\sigma$ and $(I, M, M) \not\models \varphi$. Let $(P, N)$ be the partition of $MA(\Sigma)$ induced by $(M, M)$. By Theorem 4.7, $M = \{I : I \models ob(P, N)\}$. Therefore, by Definition 4.8, $(P, N) = Prt(\Sigma, ob(P, N))$. Then, since $(I, M, M) \not\models \varphi$, by Theorem 4.10 it follows that $I \not\models \varphi(ob(P, N))$, and since $(I, M, M) \models \sigma$, by Lemma 4.6 $I \models \sigma(P, N)$, therefore $I \models \sigma(P, N) \wedge \neg\varphi(ob(P, N))$. Now suppose there exists a partition $(P', N')$ of $MA(\Sigma)$ such that $(P', N') \neq (P, N)$ and none of conditions (c1), (c2), and (c3) holds. Then, since $\sigma(P', N')$ is satisfiable, there exists an interpretation $J$ such that $J \models \sigma(P', N')$, and since $(P', N') = Prt(\Sigma, ob(P', N'), ob(P, N))$, from Lemma 4.6 it follows that there exists an interpretation $J$ such that $(J, M', M) \models \sigma$, where $M' = \{I : I \models ob(P', N')\}$. Then, since condition (c3) does not hold, the propositional formula $ob(P, N) \supset ob(P', N')$ is valid, which implies that $M' \supseteq M$. Now, if $M' = M$, then $(P', N')$ would be the partition induced by $(M, M)$, thus contradicting the hypothesis $(P', N') \neq (P, N)$. Hence, $M' \supset M$, and since $(J, M', M) \models \sigma$, it follows that $(I, M)$ is not an MBNF model for $\sigma$. Contradiction. Consequently, condition (c) in the algorithm holds, therefore MBNF-Not-Entails$(\sigma, \varphi)$ returns true.

*Only-if part.* Suppose MBNF-Not-Entails$(\sigma, \varphi)$ returns true. Then, there exists a partition $(P, N)$ of $MA(\Sigma)$ such that conditions (a), (b), and (c) hold. Let $M = \{I : I \models ob(P, N)\}$. Since $(P, N) = Prt(\Sigma, ob(P, N))$, by Definition 4.8 $(P, N)$ is the partition induced by $(M, M)$. And since $\sigma(P, N) \wedge \neg\varphi(ob(P, N))$ is satisfiable, it follows that there exists an interpretation $I$ such that $I \models \sigma(P, N)$ and $I \not\models \varphi(ob(P, N))$, hence, by Lemma 4.6, $(I, M, M) \models \sigma$ and $(I, M, M) \not\models \varphi$. Now suppose $(I, M)$ is not an MBNF model for $\sigma$. Then, there exists a set $M'$ and an interpretation $J$ such that $M' \supset M$ and $(J, M', M) \models \sigma$. Let $(P', N')$ be the partition of $MA(\Sigma)$ induced by $(M', M)$. Since $M = \{I : I \models ob(P, N)\}$, it follows that $M'$ contains at least one interpretation $J$ which does not satisfy $ob(P, N)$, and since $ob(P, N) = \bigwedge_{B\psi \in P} \psi(P, N)$, $J$ does not satisfy at least one formula of the form $\psi(P, N)$ such that $B\psi \in P$. Therefore, $P' \neq P$, which implies that $(P', N') \neq (P, N)$. Then, since $(J, M', M) \models \sigma$, by Lemma 4.6 $J \models \sigma(P', N')$, hence $\sigma(P', N')$ is satisfiable. Now let $M'' = \{I : I \models ob(P', N')\}$. By Lemma 4.14, it follows that $(P', N')$ is the partition induced by $(M'', M)$, therefore, by Definition 4.8, $(P', N') = Prt(\Sigma, ob(P', N'), ob(P, N))$. Moreover, since $M' \supset M$, it follows that the propositional formula $ob(P, N) \supset ob(P', N')$ is valid, hence the formula $ob(P, N) \wedge \neg ob(P', N')$ is unsatisfiable. Consequently, $(P', N')$ does not satisfy condition (c) in the algorithm, thus contradicting the hypothesis. Therefore, $(I, M)$ is an MBNF model for $\sigma$, and since $(I, M, M) \not\models \varphi$, it follows that $\sigma \not\models_{\text{MBNF}} \varphi$, thus proving the thesis. $\square$

We point out the fact that the algorithm MBNF-Not-Entails does not rely on a theorem prover for a modal logic: thus, "modal reasoning" is not actually needed for reasoning in MBNF. This is an interesting peculiarity that MBNF shares with other nonmonotonic modal formalisms, like autoepistemic logic (Moore, 1985) or the autoepistemic logic of knowledge (Schwarz, 1991).

### 4.3 Reasoning in Flat MBNF

We now study reasoning in flat MBNF theories. The main reason for taking into account the flat fragment of MBNF is the fact that reasoning in many of the best known nonmonotonic formalisms like default logic, circumscription, and logic programming, can be reduced to reasoning in flat MBNF theories (Lifschitz, 1994).

It is known that, if $\sigma \in \mathcal{L}_M^1$ and $\varphi \in \mathcal{L}_B^S$, then it is possible to reduce the entailment $\sigma \models_{\text{MBNF}} \varphi$ to reasoning in logic $\mathsf{S4F}_{MDD}$, by translating MBNF formulas into unimodal formulas of $\mathsf{S4F}_{MDD}$ (Schwarz & Truszczyński, 1994). Thus, the procedure for deciding entailment in the logic $\mathsf{S4F}_{MDD}$ presented by Marek and Truszczyński (1993) can be employed for computing the entailment $\sigma \models_{\text{MBNF}} \varphi$. In the following we study a more general problem, that is entailment $\sigma \models_{\text{MBNF}} \varphi$ in the case $\sigma \in \mathcal{L}_M^1$ and $\varphi \in \mathcal{L}_B$, and present a specialized algorithm for this problem, which is simpler than the more general reasoning method for $\mathsf{S4F}_{MDD}$.

In Figure 2 we report the algorithm Flat-Not-Entails for computing such an entailment. In the algorithm, $P_n$ denotes the subset of modal atoms from $P$ prefixed by the modality $not$, i.e. $P_n = \{not\, \psi : not\, \psi \in P\}$.

Informally, correctness of the algorithm Flat-Not-Entails is established by the fact that, if $\sigma \in \mathcal{L}_M^1$, then (a), (b), and (c) are necessary and sufficient conditions on a partition $(P, N)$ in order to establish whether it is induced by a pair $(M, M)$ such that there exists an MBNF model for $\sigma$ of the form $(I, M)$. In particular, condition (c) states that $B(ob(P, N))$ must be a consequence of $\sigma(P_n, N)$ in modal logic $\mathsf{S5}$,[1] since it can be shown that if $\sigma(P_n, N) \supset B(ob(P, N))$ is not valid in $\mathsf{S5}$, then the guess on the modal atoms of the form $B\varphi$ in $P$ is not minimal. We illustrate this fact through the following example.

**Example 4.16** Let

$$\sigma = (Ba \wedge not(c \vee d)) \vee (B(a \wedge b) \wedge \neg Bc) \vee Bc$$

---

1. We denote as $B$ the modal operator used in $\mathsf{S5}$.

---

**Algorithm** Flat-Not-Entails($\sigma, \varphi$)
**Input:** formula $\sigma \in \mathcal{L}_M^1$, formula $\varphi \in \mathcal{L}_B$;
**Output:** true if $\sigma \not\models_{\text{MBNF}} \varphi$, false otherwise.
**begin**
**if there exists** partition $(P, N)$ of $MA(\sigma)$
**such that**
    (a) $(P, N) = Prt(\Sigma, ob(P, N))$ **and**
    (b) $\sigma(P, N) \wedge \neg\varphi(ob(P, N))$ is satisfiable **and**
    (c) $\sigma(P_n, N) \supset B(ob(P, N))$ is valid in $\mathsf{S5}$
**then return** true
**else return** false
**end**

---

Figure 2: Algorithm Flat-Not-Entails.

and suppose

$$P = \{Ba, B(a \wedge b), not(c \vee d)\}$$
$$N = \{Bc\}$$

Then,

$$\sigma(P_n, N) = (Ba \wedge \mathsf{true}) \vee (B(a \wedge b) \wedge \neg\mathsf{false}) \vee \mathsf{false},$$

which is propositionally equivalent to $Ba \vee B(a \wedge b)$, and $ob(P, N) = a \wedge (a \wedge b)$, which is equivalent to $a \wedge b$. Now, $Ba \vee B(a \wedge b) \supset B(a \wedge b)$ is not valid in $\mathsf{S5}$, which is proved by the fact that the set of interpretations $M' = \{I : I \models a\}$ is such that $M' \models (Ba \vee B(a \wedge b)) \wedge \neg B(a \wedge b)$. Indeed, the set of interpretations $M'$ can be immediately used in order to prove that $(P, N)$ does not identify any MBNF model for $\sigma$. In fact, let $M = \{J : J \models a \wedge b\}$: it is immediate to see that, for each interpretation $I$, $(I, M', M) \models \sigma$, and since $M' \supset M$, $(I, M)$ is not an MBNF model for $\sigma$. $\square$

Finally, condition (b) corresponds to check whether there exists an interpretation $I$ satisfying $\neg\varphi(ob(P, N))$: in fact, if such an interpretation exists, then $(I, M)$ is an MBNF model for $\sigma$ which does not satisfy $\varphi$.

Therefore, the algorithms MBNF-Not-Entails and Flat-Not-Entails only differ in the way in which it is verified whether the MBNF structure associated with a partition $(P, N)$ satisfies the preference semantics provided by Definition 2.1, which is implemented through condition (c) in both algorithms. In the algorithm MBNF-Not-Entails, a partition is checked against all other partitions of $MA(\sigma)$, while in the algorithm Flat-Not-Entails it is sufficient to verify that the partition $(P, N)$ satisfies a "local" property. As shown in the next section, such a difference reflects the different computational properties of the entailment problem in the two cases.

In order to establish correctness of the algorithm, we need a preliminary lemma.

**Lemma 4.17** *Let $\sigma \in \mathcal{L}_M^1$ and let $(P, N)$ be the partition induced by a structure $(M, M)$. Then, $(I, M)$ is an MBNF model for $\sigma$ iff for each $M' \supset M$ the positive formula $\sigma(P_n, N)$ is not satisfied by $M'$.*

*Proof.* Suppose $(I, M)$ is an MBNF model for $\sigma$, and let $(P, N)$ be the partition induced by $(M, M)$. Let $M'$ be any set of interpretations such that $M' \supset M$. Then, $(M', M) \not\models \sigma$. Since $\sigma \in \mathcal{L}_M^1$ and $M' \supset M$, this implies that for each modal atom $\xi$ in $N$, $(M', M) \not\models \xi$. Moreover, for each modal atom *not* $\psi \in P$, $(M', M) \models$ *not* $\psi$. Therefore, by Lemma 4.6, $(M', M) \not\models \sigma(P_n, N)$. Now, since $\sigma \in \mathcal{L}_M^1$, $\sigma(P_n, N)$ is a flat positive formula, hence its satisfiability only depends on the structure $M'$, therefore $M' \not\models \sigma(P_n, N)$.

Conversely, suppose $(I, M)$ is not an MBNF model for $\sigma$, and let $(P, N)$ be the partition induced by $(M, M)$. Then, there exists a set of interpretations $M'$ such that $M' \supset M$ and $(M', M) \models \sigma$. As shown before, this implies that the positive formula $\sigma(P_n, N)$ is satisfied by $M'$. $\qquad\square$

As observed in Section 2, the class of universal Kripke structures characterizes modal logic S5. This immediately implies the following property.

**Lemma 4.18** *A formula $\varphi \in \mathcal{L}_B^S$ is valid in S5 iff, for each set of interpretations $M$, the formula $\neg\varphi$ is not satisfied by $M$.*

Based on the above property, we are now able to prove correctness of the algorithm Flat-Not-Entails.

**Theorem 4.19** *Let $\sigma \in \mathcal{L}_M^1$ and $\varphi \in \mathcal{L}_B$. Then, Flat-Not-Entails$(\sigma, \varphi)$ returns* true *iff $\sigma \not\models_{\text{MBNF}} \varphi$.*

*Proof.* *If-part.* If $\sigma \not\models_{\text{MBNF}} \varphi$, then there exists an MBNF model $(I, M)$ for $\sigma$ such that $(I, M, M) \not\models \varphi$. Let $(P, N)$ be the partition of $MA(\sigma)$ induced by $(M, M)$. From Theorem 4.7 it follows that $M = \{J : J \models ob(P, N)\}$. Therefore, by Definition 4.8, $(P, N) = Prt(\Sigma, ob(P, N))$, hence condition (a) in the algorithm holds.

Now let $\sigma' = \sigma(P_n, N)$, and suppose the formula $\sigma' \supset B(ob(P, N))$ is not valid in S5. Then, since the formula $\sigma' \supset B(ob(P, N))$ belongs to $\mathcal{L}_B^S$, by Lemma 4.18 it follows that there exists a set of interpretations $M'$ satisfying $\sigma' \wedge \neg B(ob(P, N))$. Let $(P', N')$ be the partition of $MA(\sigma')$ induced by $(M', M')$, and let $M'' = \{I : I \models ob(P', N')\}$. Since $ob(P, N) = \bigwedge_{B\varphi \in MA(\sigma')} \varphi$, by Definition 4.3 it follows that $ob(P, N) \supset ob(P', N')$ is a valid propositional formula, hence $M'' \supseteq M$. Now, since by hypothesis $M' \models \neg B(ob(P, N))$, it follows that $M'' \supset M$. Moreover, since $\sigma' \in \mathcal{L}_B$, by Lemma 4.14 it follows that $(P', N')$ is the partition induced by $(M'', M'')$, and since $M' \models \sigma'$ and $\sigma'$ is flat, $\sigma'(P', N')$ is equivalent to true, therefore $M'' \models \sigma'(P', N')$ and, by Lemma 4.6, $M'' \models \sigma'$. On the other hand, since $M'' \supset M$, by Lemma 4.17 it follows that $M'' \not\models \sigma'$. Contradiction. Hence, $\sigma' \supset B(ob(P, N))$ is valid in S5, consequently condition (c) of the algorithm holds.

Finally, since $(I, M, M) \not\models \varphi$ and $M = \{J : J \models ob(P, N)\}$, by Theorem 4.10 it follows that $I \not\models \varphi(ob(P, N))$. Moreover, since $(I, M, M) \models \sigma$, from Lemma 4.6 it follows that $I \models \sigma(P, N)$, consequently $I \models \sigma(P, N) \wedge \neg\varphi(ob(P, N))$, hence the propositional formula $\sigma(P, N) \wedge \neg\varphi(ob(P, N))$ is satisfiable. Therefore, conditions (a), (b), and (c) in the algorithm hold, which implies that Flat-Not-Entails$(\sigma, \varphi)$ returns true.

*Only-if-part.* If Flat-Not-Entails$(\sigma, \varphi)$ returns true, then there exists a partition $(P, N)$ of $MA(\sigma)$ for which conditions (a), (b), and (c) of the algorithm hold. Let $M = \{J : J \models ob(P, N)\}$. By Definition 4.8, $(P, N)$ is the partition of $MA(\sigma)$ induced by $(M, M)$. Now, since $\sigma(P, N) \wedge \neg\varphi(ob(P, N))$ is satisfiable, there exists an interpretation $I$ such that $I \models \sigma(P, N)$ and $I \models \neg\varphi(ob(P, N))$, hence by Lemma 4.6 $(I, M, M) \models \sigma$, and by Theorem 4.10 $(I, M, M) \not\models \varphi$, therefore we only have to show that $(I, M)$ is an MBNF model for $\sigma$. So, let us suppose $(I, M)$ is not an MBNF model for $\sigma$. Then, by Lemma 4.17 there exists $M' \supset M$ such that $\sigma(P_n, N)$ is satisfied in $M'$. Now, condition (c) in the algorithm implies that $B(ob(P, N))$ is a consequence of $\sigma(P_n, N)$ in S5, therefore $ob(P, N)$ is satisfied by each interpretation in $M'$, that is, $M' \subseteq \{J : J \models ob(P, N)\}$, which contradicts the hypothesis $M' \supset M = \{J : J \models ob(P, N)\}$. Consequently, $(I, M)$ is an MBNF model for $\sigma$. $\square$

We remark the fact that the algorithm Flat-Not-Entails can be seen as a generalization of known methods for query answering in Reiter's default logic, Moore's autoepistemic logic, and (disjunctive) logic programming under the stable model (and answer set) semantics. In particular, condition (c) in the algorithm can be seen as a generalization of the minimality check used in (disjunctive) logic programming for verifying stability of a model of a logic program (Gelfond & Lifschitz, 1990, 1991).

## 5. Complexity Results

In this section we provide a computational characterization of reasoning in MBNF.

We first briefly recall the complexity classes in the *polynomial hierarchy*, and refer to (Johnson, 1990; Papadimitriou, 1994) for further details about the complexity classes mentioned in the paper. $P^A$ ($NP^A$) is the class of problems that are solved in polynomial time by deterministic (nondeterministic) Turing machines using an oracle for $A$ (i.e. that solves in constant time any problem in $A$). The classes $\Sigma_k^p$, $\Pi_k^p$ and $\Delta_k^p$ of the polynomial hierarchy are defined by $\Sigma_0^p = \Pi_0^p = \Delta_0^p = P$, and for $k \geq 0$, $\Sigma_{k+1}^p = NP^{\Sigma_k^p}$, $\Pi_{k+1}^p = co\Sigma_{k+1}^p$ and $\Delta_{k+1}^p = P^{\Sigma_k^p}$. In particular, the complexity class $\Sigma_2^p$ is the class of problems that are solved in polynomial time by a nondeterministic Turing machine that uses an NP-oracle, and $\Pi_2^p$ is the class of problems that are complement of a problem in $\Sigma_2^p$, while $\Sigma_3^p$ is the class of problems that are solved in polynomial time by a nondeterministic Turing machine that uses an $\Sigma_2^p$-oracle, and $\Pi_3^p$ is the class of problems that are complement of a problem in $\Sigma_3^p$. It is generally assumed that the polynomial hierarchy does not collapse: hence, a problem in the class $\Sigma_2^p$ or $\Pi_2^p$ is considered computationally easier than a $\Sigma_3^p$-hard or $\Pi_3^p$-hard problem.

As for the complexity of entailment in MBNF, we start by establishing a lower bound for reasoning in propositional MBNF theories. To this end, we exploit the correspondence between MBNF and the logic of minimal knowledge $S5_G$ (Halpern & Moses, 1985). Indeed, as stated by Proposition 2.5, there is a one-to-one correspondence between MBNF models and $S5_G$ models of positive subjective theories.

**Lemma 5.1** *Let $\sigma \in \mathcal{L}_M^S$ and let $\varphi \in \mathcal{L}_B$. Then, the problem of deciding whether $\sigma \models_{\text{MBNF}} \varphi$ is $\Pi_3^p$-hard.*

*Proof.* As shown by (Donini et al., 1997b), entailment in $\mathsf{S5}_G$ is $\Pi_3^p$-complete. Therefore, by Proposition 2.5, for subjective (and hence for general) MBNF theories, entailment is $\Pi_3^p$-hard. □

Then, we show that the entailment problem in propositional MBNF is complete with respect to the class $\Pi_3^p$.

**Theorem 5.2** *Let* $\sigma \in \mathcal{L}_M$ *and let* $\varphi \in \mathcal{L}_B$. *Then, the problem of deciding whether* $\sigma \models_{\text{MBNF}} \varphi$ *is* $\Pi_3^p$-complete.

*Proof.* Hardness with respect to $\Pi_3^p$ follows from Lemma 5.1. As for membership in $\Pi_3^p$, we analyze the complexity of the algorithm MBNF-Not-Entails reported in Figure 1. In particular, observe that:

- given $(P, N)$, the formula $ob(P, N)$ can be computed in polynomial time with respect to the size of $P$. Moreover, by Lemma 4.9 it follows that, since $MA(\sigma)$ has size linear with respect to the size of $\sigma$, construction of the partition $Prt(\Sigma, ob(P, N))$ can be performed through a linear number (with respect to the size of $\sigma$) of calls to an NP-oracle for propositional satisfiability. Therefore, condition (a) can be checked through a linear number (in the size of the input) of calls to an NP-oracle;

- since $\varphi(ob(P, N)) = \varphi(Prt(\varphi, ob(P, N)))$, the formula $\neg\varphi(ob(P, N))$ can be computed in time linear with respect to the size of $\varphi \wedge ob(P, N)$ using an NP-oracle. And since, given $\sigma$ and $(P, N)$, $\sigma(P, N)$ can be computed in polynomial time with respect to the size of the input, it follows that condition (b) can be computed through a linear number (in the size of the input) of calls to an NP-oracle;

- given a partition $(P', N')$, each of the conditions (c1), (c2) and (c3) (analogous to conditions (a) and (b)) can be checked in polynomial time, with respect to the size of $\sigma$, using an NP-oracle. Therefore, since the guess of the partition $(P', N')$ of $MA(\sigma)$ requires a nondeterministic choice, falsity of condition (c) can be decided in $\Sigma_2^p$, which implies that verifying whether condition (c) holds can be decided in $\Pi_2^p$.

Since the guess of the partition $(P, N)$ of $MA(\sigma)$ requires a nondeterministic choice, it follows that the algorithm MBNF-Not-Entails, if considered as a nondeterministic procedure, decides $\sigma \not\models_{\text{MBNF}} \varphi$ in nondeterministic polynomial time (with respect to the size of $\sigma \wedge \varphi$), using a $\Sigma_2^p$-oracle. Thus, we obtain an upper bound of $\Sigma_3^p$ for the non-entailment problem, which implies that entailment in MBNF is in $\Pi_3^p$. □

The previous analysis also allows for a computational characterization of the logic MKNF (Lifschitz, 1991), which is a slight modification of MBNF. Indeed, it is known (Lifschitz, 1994) that, for each theory $\Sigma \subseteq \mathcal{L}_M$, $M$ is an MKNF model of $\Sigma$ iff, for each interpretation $I$, $(I, M)$ is an MBNF model of the subjective theory $\Sigma' = \{B\varphi : \varphi \in \Sigma\}$. Therefore, from Proposition 2.5 and from $\Pi_3^p$-hardness of entailment in $\mathsf{S5}_G$ (Donini et al., 1997b), it follows that entailment in MKNF is $\Pi_3^p$-hard. Then, since $\Sigma \models_{\text{MKNF}} \varphi$ iff $\Sigma' \models_{\text{MBNF}} B\varphi$ (Lifschitz, 1994), it follows that entailment in MKNF can be polynomially reduced to entailment in MBNF, hence such a problem belongs to $\Pi_3^p$. Therefore, the following property holds.

**Theorem 5.3** *Entailment in propositional* MKNF *is* $\Pi_3^p$-complete.

Finally, the previous theorem provides a computational characterization of the logic of grounded knowledge and justified assumptions GK (Lin & Shoham, 1992). In fact, the logic GK can be considered as a syntactic variant of the propositional fragment of MKNF. Therefore, skeptical entailment in GK is $\Pi_3^p$-complete.

**Remark.** The computational properties of MBNF and its variants relate such formalisms to ground nonmonotonic modal logics (Eiter & Gottlob, 1992; Donini et al., 1997b; Rosati, 1999). Notably, ground nonmonotonic modal logics share with MBNF the interpretation in terms of minimal knowledge (or minimal belief) of the modality $B$; specifically, as already mentioned, the propositional fragment of MBNF can be considered as built upon $\mathsf{S5}_G$ by adding a second modality *not*. Therefore, it turns out that, in the propositional case, adding a "negation by default" modality to the $\mathsf{S5}$ logic of minimal knowledge does not increase the computational complexity of reasoning, while adding a minimal knowledge modality to AEL does increase the complexity of deduction. We can thus summarize as follows: minimal knowledge is computationally harder than negation as failure. □

We now study the complexity of entailment for flat MBNF theories. First, it is known that, in the case of flat MBNF theories and *subjective* queries, entailment is $\Pi_2^p$-complete: membership in the class $\Pi_2^p$ is a consequence of the fact that flat MKNF theories can be polynomially embedded into McDermott and Doyle's nonmonotonic modal logic $\mathsf{S4F}$ (Schwarz & Truszczyński, 1994, Proposition 3.2), whose entailment problem is $\Pi_2^p$-complete (Marek & Truszczyński, 1993), while $\Pi_2^p$-hardness follows from the existence of a polynomial-time embedding of propositional default theories into flat MBNF theories (Lifschitz, 1994). Therefore, the following property holds.

**Proposition 5.4** *Let* $\sigma \in \mathcal{L}_M^1$ *and let* $\varphi \in \mathcal{L}_B^S$. *Then, the problem of deciding whether* $\sigma \models_{\mathrm{MBNF}} \varphi$ *is* $\Pi_2^p$*-complete.*

As for complexity of entailment of generic queries with respect to flat MBNF theories, we analyze the complexity of the algorithm Flat-Not-Entails reported in Figure 2. As shown before, both condition (a) and condition (b) can be checked through a linear number (with respect to the size of the input) of calls to an NP-oracle. Moreover, validity in modal logic $\mathsf{S5}$ is a coNP-complete problem (Halpern & Moses, 1992). Hence, each of the conditions in the algorithm can be computed through a number of calls to an oracle for the propositional validity problem which is polynomial in the size of the input, and since the guess of the partition $(P, N)$ of $MA(\sigma)$ requires a nondeterministic choice, it follows that the algorithm runs in $\Sigma_2^p$. Therefore, the following property holds.

**Theorem 5.5** *Let* $\sigma \in \mathcal{L}_M^1$ *and let* $\varphi \in \mathcal{L}_B$. *Then, the problem of deciding whether* $\sigma \models_{\mathrm{MBNF}} \varphi$ *is* $\Pi_2^p$*-complete.*

*Proof.* Membership of the problem to the class $\Pi_2^p$ is implied by the algorithm Flat-not-entails, whereas $\Pi_2^p$-hardness is implied by Proposition 5.4. □

Hence, the algorithm Flat-Not-Entails is "optimal" in the sense that it matches the lower bound for the entailment problem.

Finally, we remark that the subset of flat MBNF theories in conjunctive normal form can be seen as a further extension of the framework of *generalized logic programming* introduced by Inoue and Sakama (1994), which in turn is an extension of the disjunctive logic programming framework under the stable model semantics (Gelfond & Lifschitz, 1991). Roughly speaking, flat MBNF theories in conjunctive normal form correspond to rules of generalized logic programs in which propositional formulas (instead of literals) are allowed as goals. The above computational characterization implies that such an extension of the framework of logic programming under the stable model semantics does not affect the worst-case complexity of the entailment problem, which is $\Pi_2^p$-complete just like entailment in logic programs with disjunction under the stable model semantics (Eiter & Gottlob, 1995). Such a result extends analogous properties (Marek, Truszczyński, & Rajasekar, 1995) to the case of disjunctive logic programs.

## 6. Conclusions

In this paper we have investigated the problem of reasoning in the propositional fragment of MBNF. The main results presented can be summarized as follows:

- the negation as failure modality *not* of MBNF exactly corresponds to negative introspection in AEL. This implies that the logic MBNF can be viewed as the conservative extension of two different nonmonotonic modal logics: Halpern and Moses' logic of minimal knowledge $\mathsf{S5}_G$ and Moore's AEL;

- reasoning in the propositional fragment of MBNF lies at the third level of the polynomial hierarchy, hence (unless the polynomial hierarchy does not collapse) reasoning in MBNF is harder than reasoning in the best known propositional nonmonotonic logics, like default logic, autoepistemic logic, and circumscription;

- we have defined methods for reasoning in MBNF, which subsume and generalize well-known nonmonotonic reasoning algorithms used in logic programming (Gelfond & Lifschitz, 1991), default logic (Gottlob, 1992), and autoepistemic logic (Marek & Truszczyński, 1993);

- we have studied the flat fragment of MBNF and its relationship with the logic programming paradigm.

As for the computational aspects of reasoning in MBNF, the results presented in Section 5 prove that one source of complexity is due to the presence of nested occurrences of modalities in the theory, since reasoning in flat MBNF is computationally easier than in the general case.

It can be proven that another source of complexity lies in the underlying objective language. In fact, if we consider $\mathcal{L}'$ to be a tractable fragment of propositional logic, then the complexity of reasoning in the modal language $\mathcal{L}'_M$ built upon $\mathcal{L}'$ is lower than in the general case. In particular, it is easy to see that, under the assumption that entailment in $\mathcal{L}'$ can be computed in polynomial time, the algorithm MBNF-Not-Entails provides an upper bound of $\Pi_2^p$ for MBNF-entailment in the fragment $\mathcal{L}'_M$.

One possible development of the present work is towards the analysis of reasoning about minimal belief and negation as failure in a first-order setting: in particular, it should be interesting to see whether it is possible to extend the techniques developed for the propositional case to a more expressive language. A first attempt in this direction is reported by Donini et al. (1997a).

## Acknowledgments

This research has been partially supported by Consiglio Nazionale delle Ricerche, grant 203.15.10.